\documentclass[sigconf]{acmart}
 \usepackage{textcomp, libertine}
\usepackage{booktabs} 
\usepackage[normalem]{ulem}
\usepackage{multirow}
\usepackage{bm}
\useunder{\uline}{\ul}{}

\usepackage{subcaption}
\usepackage[para]{footmisc}
\usepackage{enumitem}
\usepackage[ruled]{algorithm2e}
\usepackage{dsfont}
\usepackage{pifont}
\hypersetup{
,bookmarksnumbered=true%
,hypertexnames=false%
,breaklinks=true%
,colorlinks=true%
,linkcolor=black%
,citecolor=black%
,urlcolor=black%
}

\definecolor{tealblue}{rgb}{0.21, 0.46, 0.53}
\definecolor{wildstrawberry}{rgb}{1.0, 0.26, 0.64}
\definecolor{ao(english)}{rgb}{0.0, 0.5, 0.0}

\def\ignore#1{}

\copyrightyear{2021}
\acmYear{2021}
\setcopyright{rightsretained}
\acmConference[CIKM '21]{Proceedings of the 30th ACM International Conference on Information and Knowledge Management}{November 1--5, 2021}{Virtual Event, QLD, Australia}
\acmBooktitle{Proceedings of the 30th ACM International Conference on Information and Knowledge Management (CIKM '21), November 1--5, 2021, Virtual Event, QLD, Australia}
\acmDOI{10.1145/3459637.3482281}
\acmISBN{978-1-4503-8446-9/21/11}

\begin{document}

\settopmatter{printacmref=true}

\fancyhead{}




\title{Natural Language Understanding with Privacy-Preserving BERT}

\author{Chen Qu, Weize Kong, Liu Yang, Mingyang Zhang, Michael Bendersky, Marc Najork}
\thanks{Work done during Chen's internship at Google.}

\affiliation{%
	\institution{
		Google} 
	\city{Mountain View} 
	\state{CA} 
	\postcode{94043}
	\country{United States}
}
\email{{cqu,weize,yangliuy,mingyang,bemike,najork}@google.com}

\begin{abstract}
Privacy preservation remains a key challenge in data mining and Natural Language Understanding (NLU). Previous research
shows that the input text or even text embeddings can leak private information. This concern motivates our research on effective privacy preservation approaches for pretrained Language Models (LMs). We investigate the privacy and utility implications of applying $d\chi$-privacy, a variant of Local Differential Privacy, to BERT fine-tuning in NLU applications. More importantly, we further propose privacy-adaptive LM pretraining methods and show that our approach can boost the utility of BERT dramatically while retaining the same level of privacy protection. We also quantify the level of privacy preservation and provide guidance on privacy configuration. Our experiments and findings lay the groundwork for future explorations of privacy-preserving NLU with pretrained LMs.

\end{abstract}


\maketitle


\section{Introduction}
\label{sec:intro}
The recent development of Deep Learning (DL) has led to notable success in Natural Language Understanding (NLU). Data-driven neural models are being applied to a rich variety of NLU applications, such as sentiment analysis~\cite{sst}, question answering~\cite{Qu2019BERTWH}, information retrieval~\cite{Zamani2018FromNR}, and text generation~\cite{Brown2020LanguageMA}. Many of these technologies have been deployed on the cloud by industrial service providers to process user data from personal customers, small businesses, and large enterprises. 
However, the rapid growth of NLU technologies also comes with a series of privacy challenges due to the sensitive nature of user data. In NLU, the input text or even the text vector representations can leak private information or even identify specific authors ~\cite{Pan2020PrivacyRO,Coavoux2018PrivacypreservingNR,Li2018TowardsRA,Song2020InformationLI}. 
This lack of privacy guarantees may impede privacy-conscious users from releasing their data to service providers.
Thus, service providers may suffer from the deficiency of genuine and evolving user data to train and evaluate NLU models.
Besides, unintended data disclosure and other privacy breaches may result in litigation, fines, and reputation damages for service providers. 
These concerns necessitate our research on privacy-preserving NLU. 

Specifically, we identify two challenges for privacy-preserving NLU. The first challenge is how to privatize users' text data in a Local Privacy setting, i.e., anonymize text to prevent leakage of private information. Prior work has applied Differential Privacy (DP)~\cite{Dwork2006CalibratingNT} and its variants to address similar privatization issues -- originally for statistical databases~\cite{Dwork2006CalibratingNT} and more recently for DL~\cite{Abadi2016DeepLW, Shokri2015PrivacypreservingDL} and NLU~\cite{Lyu2020TowardsDP, Feyisetan2020PrivacyPreservingTA, Lyu2020DifferentiallyPR}. 
However, in the context of NLU, many previous works mostly focus on a \textit{Centralized Privacy} setting, which assumes a trusted centralized data aggregator to collect and process users' text data for training NLU models~\cite{McMahan2018LearningDP}. 
This solution, however, might not be sufficient for many users who are concerned with their sensitive or proprietary text data, when used for model training and serving.
Thus, text privatization without a trusted data aggregator, also referred to as a \textit{Local Privacy} setting, has become a pressing problem that remains less explored.

To tackle this challenge,
we consider \textit{Local Differential Privacy} (LDP) as the backbone of our privacy-preserving mechanism. In this setting, users perturb each individual data entry to provide plausible deniability~\cite{Bindschaedler2017PlausibleDF} with respect to the original input \textit{before} releasing it to the service providers. 
LDP also has advantages over federated learning~\cite{McMahan2017CommunicationEfficientLO} as discussed in Sec.~\ref{sec:related-work}.
Specifically, we adopt a text privatization mechanism recently proposed by \citet{Feyisetan2020PrivacyPreservingTA}. This mechanism is based on $d\chi$-privacy (Sec.~\ref{subsec:preliminaries}). It relaxes LDP to preserve more information from the input so that it is more practical for NLU applications~\cite{Feyisetan2020PrivacyPreservingTA}.

\begin{figure}[t]
    \centering
    \includegraphics[width=0.4\textwidth]{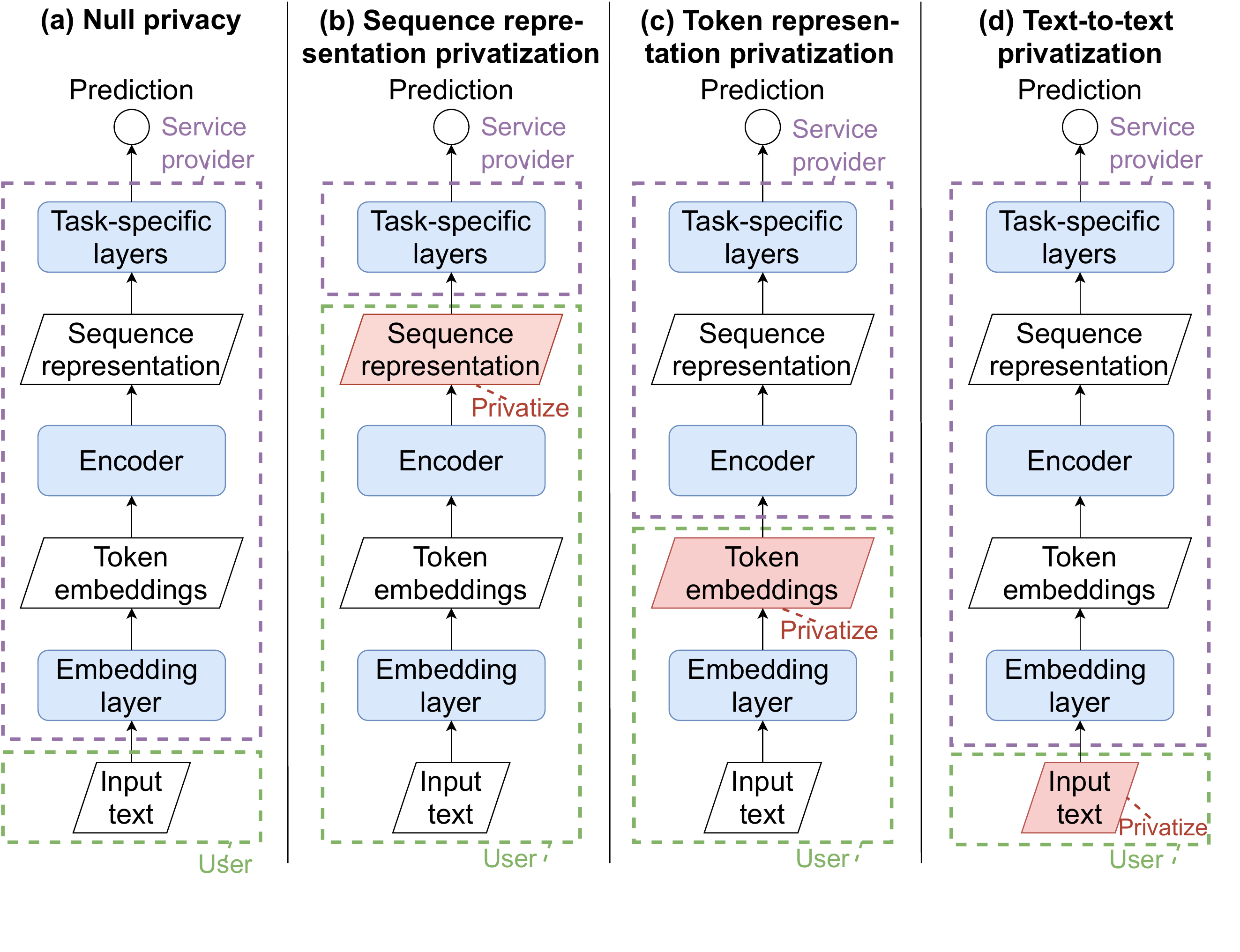}
    \vspace{-0.8cm}
    \caption{Illustrations of different privacy-constrained methods for training a typical NLU model.}
    \label{fig:fine-tuning}
\end{figure}

The second challenge, which is also the focus of this work, is how to improve the utility of NLU models under Local Privacy settings, where the text is already privatized before model training and serving. Recent progress of pretrained Language Models (LMs) has led to great success in NLU. However, to the best of our knowledge, these research questions have not been well studied: 
(a) Can pretrained LMs adapt to privatized text input? 
(b) What is the most practical way to apply text privatization for pretrained LMs so that we retain the most utility (Fig.~\ref{fig:fine-tuning})?
(c) Can we improve pretrained LMs to adapt to the privatized input via pretraining? 

To answer these research questions, we first systematically discuss three privacy-constrained fine-tuning methods that apply text privatization at different stages of the NLU model: sequence representations (Fig.~\ref{fig:fine-tuning}.b), token representations (Fig.~\ref{fig:fine-tuning}.c), and input text (Fig.~\ref{fig:fine-tuning}.d).
Along with previous research~\cite{Feyisetan2020PrivacyPreservingTA,Lyu2020DifferentiallyPR}, our work is another fundamental step to paint a more complete picture of text privatization for NLU.
\citet{Lyu2020DifferentiallyPR} only looked at sequence representation privatization in BERT fine-tuning. This method does not address training-time privacy issues and incurs heavy computational costs for users (Sec.~\ref{sec:related-work}). In contrast, our work guarantees tunable privacy protection (Sec.~\ref{sec:privacy-experiments}) at both training and inference time by focusing on the token-level privatization (Fig.~\ref{fig:fine-tuning}.c/d). 
\citet{Feyisetan2020PrivacyPreservingTA} only considered text-to-text privatization while we also study token representation privatization. Furthermore, we investigate how to improve NLU performance on top of the privacy mechanisms.

More importantly, to the best of our knowledge, our work is the first to study privacy-adaptive LM pretraining to improve the effectiveness and robustness of pretrained LMs on privatized text. 
We focus on BERT~\cite{bert} since it is one of the most widely-used pretrained LMs. 
Our privacy-adaptive pretraining approaches are based on several variants of the Masked Language Model loss we designed, to leverage large-scale public text corpora for self-supervised learning in a privacy-adaptive manner, and to address the deficiency of labeled user data in privacy-constrained fine-tuning. 

We conduct both privacy and utility experiments on two benchmark datasets. For the privacy experiments, we demonstrate and interpret the level of privacy protection by analyzing the plausible deniability statistics on the vocabulary level, as well as investigating the performance of token embedding inversion attack on actual corpora.
Based on these results, we discuss and compare two principled approaches to guide the selection of the privacy parameter. We also reveal the geometry properties of the BERT embedding space to better understand the privatization process. 

For the utility experiments, we investigate the performance of token-level privacy-constrained fine-tuning and privacy-adaptive pretraining methods.
In the fine-tuning experiments, 
we discover that the text-to-text privatization method can often improve over token representation privatization thanks to the post-processing step of nearest neighbor search.
More importantly, in pretraining experiments, we first show that BERT is able to adapt to privatization to some extent by being pretrained on fully privatized corpora. We further demonstrate that the integration of a denoising heuristic can make BERT even more robust in handling privatized representations. Another exciting finding is that token representation privatization outperforms text-to-text privatization when noise is large with privacy-adaptive pretraining. In other words, the adaption resulting from privacy-adaptive pretraining can work better than nearest neighbor search in this scenario. These results show that our privacy-adaptive pretraining approaches can make BERT more effective and robust in handling privatized text input.


\section{Related Work}
\label{sec:related-work}

\textbf{Differential Privacy}.
Differential Privacy (DP)~\cite{Dwork2006CalibratingNT} was originated from the field of statistical databases. It is one of the primary methods for defining privacy and preventing privacy breaches. At a high level, a randomized algorithm is differentially private if its output distribution is similar when the algorithm runs on two input datasets that differ in at most one data entry. Therefore, an observer seeing the output cannot tell if a particular data entry was used in the computation. 
Two settings are typically considered for DP: Centralized DP (CDP) and Local DP (LDP).
CDP assumes a trusted data collector who can collect and access the raw user data. The randomized algorithm is applied on the collected dataset to produce differentially private output for downstream use. LDP~\cite{Kasiviswanathan2008WhatCW} does not assume such a trusted data collector (e.g., users do not want service providers to access and collect their raw text messages). Instead, a randomized algorithm is applied on each individual data entry to provide plausible deniability~\cite{Bindschaedler2017PlausibleDF} before sending it to the untrusted data collector (e.g., each user privatizes the text messages before uploading them to the service providers). 

\textbf{Privacy-Preserving Deep Learning}.
Another line of research aims to train privacy-preserving DL models~\cite{Abadi2016DeepLW,McMahan2018LearningDP} by using differentially private stochastic gradient descent. The goal of those papers is to prevent the DL model from memorizing and leaking sensitive information in the training data. In contrast, some other work in this line aims to prevent an attacker from recovering information about the input text at inference time. For example, \citet{Coavoux2018PrivacypreservingNR} and \citet{Li2018TowardsRA} proposed to train deep models with adversarial learning, so that the model does not memorize unintended information. Both works provide only empirical improvements in privacy, without mathematically-sound privacy guarantees. 
Different from this work, PixelDP~\cite{Lcuyer2019CertifiedRT} applied DP to computer vision models so that the models are robust to adversarial examples.
Another important line of work studied federated learning~\cite{McMahan2017CommunicationEfficientLO} to conduct decentralized training with the user data left on users' local/mobile devices. 
However, such a learning schema is hindered by low computational resources on users' device, and it has been argued to have privacy issues \cite{Lyu2020ThreatsTF, Lyu2020TowardsFA}.
Compared with federated learning, our privacy-preserving NLU methods do not have the bottleneck caused by the lack of computational resources on users' local devices. 
Also, our approaches enjoy the flexibility of owning a privatized version of user data and thus can exploit the data for training new models for the same or new NLU tasks.

\textbf{Anonymization for NLU}.
We categorize anonymization methods used in NLU as text-to-text privatization and text representation privatization.
Typical text-to-text privatization methods include de-identification and $k$-anonymization~\cite{Bendersky2017LearningFU, Li2019MultiviewES}. The former redacts Personally Identifiable Information (PII) in the text while the latter retains only words or n-grams used by a sufficiently large number ($k$) of users, without word/n-gram sequence information. The downside is that de-identification could easily leak other sensitive information~\cite{Anandan2012tPlausibilityGW} and $k$-anonymization has the same flaw when the adversary has background knowledge. 
\citet{Feyisetan2020PrivacyPreservingTA} recently proposed a text-to-text privatization method based on DP ($d\chi$-privacy~\cite{Dwork2006CalibratingNT} to be exact). This method replaces the words in the original text with other words that are close in the embedding space in a local differential private manner. We also adopt this strategy since $d\chi$-privacy provides mathematically provable privacy guarantees regardless of the background knowledge an adversary might use. Different from that work, we study the impact of privatization on BERT and propose privacy-adaptive pretraining methods to improve its utility while maintaining the same privacy guarantees.

On the other hand, text representation privatization aims to privatize the float-vector representations for text, including term-frequency vectors, word2vec embeddings~\cite{Mikolov2013DistributedRO}, and representations from BERT~\cite{bert}. Recently, \citet{Weggenmann2018SynTFSA} used a similar DP-based method as \citet{Feyisetan2020PrivacyPreservingTA}, but with a different goal, i.e., to anonymize the term frequency vector representation of a document. Given the recent advance in representation learning with pretrained LMs, how to privatize representations from these models has become an increasingly important research problem. 
For example, \citet{Lyu2020TowardsDP} followed the idea of Unary Encoding~\cite{Wang2017LocallyDP}, and proposed a mechanism to anonymize text representation that provides $\varepsilon$-LDP. In addition, \citet{Bhowmick2018ProtectionAR} proposed a LDP mechanism to privatize high dimensional vectors, which is applied to the gradients for federated learning. This can be potentially applied to text representations. Also, \citet{Lyu2020DifferentiallyPR} looked at sequence representation privatization in BERT fine-tuning. A major drawback of their approach is that they did not address privacy issues at the training time since it requires service providers to have access to the raw user data to fine-tune the user-side encoder. This access violates the Local Privacy requirement at the training time. Their method also incurs heavy computational costs for users because the entire encoder is deployed on the user side. In comparison, our approach guarantees privacy protection at both training and inference time and incurs less  computational cost for users.

\section{Our Approach}
\label{sec:our-approach}

\subsection{Overview} 
\label{subsec:overview}
\begin{figure}[t]
    \centering
    \includegraphics[width=0.4\textwidth]{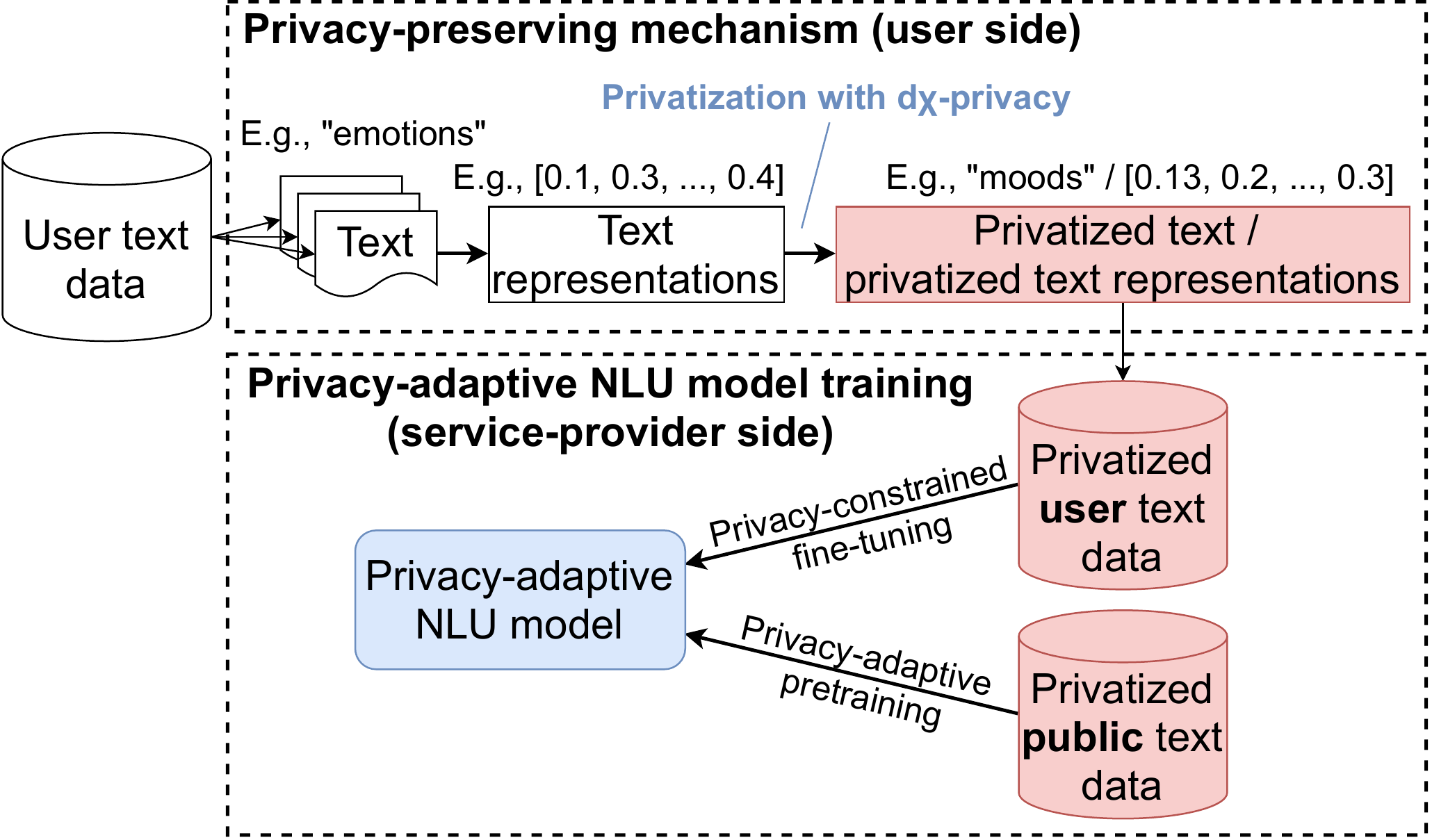}
    \vspace{-0.4cm}
    \caption{Overview.
    Users privatize input text locally. Then service providers conduct privacy-adaptive model training.
    }
    \label{fig:overview}
\end{figure}

We present an overview of our approach for privacy-preserving NLU in Fig.~\ref{fig:overview}. It contains two major stages. The first stage is the privacy-preserving mechanism (Sec.~\ref{subsec:privacy-mechanism}), where each user applies this mechanism to transform raw text to either privatized text or privatized text representations on their own local devices, and then submits the output to the service provider. The second stage is the privacy-adaptive NLU model training, where the service provider can only access the privatized text data for building NLU models. To improve the NLU performance on privatized text data, we consider various privacy-constrained fine-tuning methods (Sec.~\ref{subsec:approach-fine-tuning}) and propose privacy-adaptive pretraining methods (Sec.~\ref{subsec:approach-pretraining}). The latter further improves model utility by leveraging large-scale public text corpora that we privatized. 
We study the widely-used BERT to exemplify how our approach applies to pretrained LMs. 

\subsection{Privacy-Preserving Mechanism}
\label{subsec:privacy-mechanism}
We describe the privacy mechanism following \citet{Feyisetan2020PrivacyPreservingTA} and illustrate how it is applied to text privatization.   

\subsubsection{\textbf{Preliminaries}} 
\label{subsec:preliminaries}
We first briefly review two variants of DP related to our work -- Local Differential Privacy (LDP) and $d\chi$-privacy. Both variants are designed for the Local Privacy setting, where users need to privatize each data instance before releasing it to the untrusted data collector. Both variants employ a randomized mechanism $M: \mathcal{X}\rightarrow\mathcal{Y}$ that takes in a single data instance $x\in\mathcal{X}$ and outputs a randomized output $y\in\mathcal{Y}$.

\textbf{LDP}~\cite{Kasiviswanathan2008WhatCW} requires $M$ to satisfy, for any two inputs $x,x'\in\mathcal{X}$, 
\begin{equation}\label{eqn:ldp}
\footnotesize
\begin{aligned}
\frac{\Pr[M(x) = y]}{\Pr[M(x') = y]} \leq e^{\varepsilon}, \forall y \in \mathcal{Y},
\end{aligned}
\end{equation}
where $\varepsilon \geq 0$ is a privacy parameter. 
Intuitively, Eq.~\ref{eqn:ldp} suggests the output of $M(x)$ and $M(x')$ have very similar distributions such that an adversary cannot tell whether the input is $x$ or $x'$. In other words, $M$ provides plausible deniability~\cite{Bindschaedler2017PlausibleDF} with respect to the original input. However, LDP is a very strong privacy standard. Regardless of how unrelated $x$ and $x'$ are, LDP requires them to have similar and indistinguishable output distributions. As a result, the output may not preserve enough information from the original input and thus may hurt the utility of downstream tasks.

\label{subsubsec:dx-privacy}
\textbf{$\bm{d\chi}$-privacy~\cite{Chatzikokolakis2013BroadeningTS}}, a relaxation of LDP, was introduced to address the problem mentioned above. 
More formally, $d\chi$-privacy requires, for any two inputs $x,x'\in\mathcal{X}$, 
\begin{equation}\label{eqn:dx}
\footnotesize
\begin{aligned}
\frac{\Pr[M(x) = y]}{\Pr[M(x') = y]} \leq e^{\eta d(x, x')}, \forall y \in \mathcal{Y},
\end{aligned}
\end{equation}
where $d(x,x')$ is a distance/metric function (e.g., Euclidean distance) and $\eta\geq 0$ is the privacy parameter.\footnote{Different from the prior work~\cite{Feyisetan2020PrivacyPreservingTA}, we use $\eta$ to denote the privacy parameter in $d\chi$-privacy instead of $\varepsilon$, in order to avoid the confusion with $\varepsilon$ used in LDP and DP.} Compared with LDP, $d\chi$-privacy allows the indistinguishability of the output distributions to be scaled by the distance between the respective inputs, i.e., $\varepsilon$ in Eq.~\ref{eqn:ldp} becomes  $\eta d(x, x')$ in Eq.~\ref{eqn:dx}. This allows $M$ to produce more similar output for similar $x$ and $x'$ measured by $d(x, x')$, and thus could preserve more information from the input. That said, we should highlight that the semantics of $\eta$ in $d\chi$-privacy depends on the choice of $d$, and thus one needs to understand the structure of the underlying metric $d$ in order to interpret the privacy consequences. This necessitates our study to measure and calibrate the level of privacy protection for our case in Sec.~\ref{subsec:plausible_deniability_statistics} and \ref{subsec:token-embedding-inversion}.
Given the advantage of $d\chi$-privacy and inspired by \citet{Feyisetan2020PrivacyPreservingTA}, we adopt $d\chi$-privacy to perform token representation privatization and text-to-text privatization in Sec.~\ref{subsubsec:rep_privatization} and \ref{subsubsec:txt_privatization}. We refer our readers to \citet{Feyisetan2020PrivacyPreservingTA} for the complete privacy proof.

\subsubsection{\textbf{Threat Model}} \label{subsubsec:threat-model}
Following \citet{Feyisetan2020PrivacyPreservingTA}, we consider a threat model where each user submits a token to the service provider to conduct various downstream tasks. A user's token either appears in its clear form or in the form of a token representation. We expand users' input from a token to a text sequence in Sec.~\ref{subsubsec:seq-input}. 

\subsubsection{\textbf{Token Representation Privatization}} \label{subsubsec:rep_privatization}
It is the \textit{de facto} method for deep NLU models to represent input tokens with distributed dense vectors (e.g., word embeddings). Such token representations are typically produced by an embedding model.
Without loss of generality, a token can be a character~\cite{Zhang2015CharacterlevelCN}, a subword or wordpiece~\cite{Wu2016GooglesNM}, a word, or an n-gram~\cite{Zhao2017Ngram2vecLI}. 

To prevent the leakage of sensitive information, we adopt $d\chi$-privacy (Sec.~\ref{subsec:preliminaries}) to privatize such token representations. In this case, the input to the randomized mechanism $M$ becomes a token embedding $x\in\mathbb{R}^n$ and the output becomes $y\in\mathbb{R}^n$. For simplicity, we only consider the case where $x,y$ are of the same dimension, $n$.
Then $d\chi$-privacy can be achieved for the choice of Euclidean distance $d(x,x')=||x-x'||$ by adding random noise $N$ drawn from an $n$-dimensional distribution with density $p(N) \propto \exp(- \eta ||N||)$~\cite{Dwork2006CalibratingNT},
\begin{equation}\label{eqn:dx-N}
\footnotesize
\begin{aligned}
M_\text{rep}(x) = x + N,
\end{aligned}
\end{equation}
where we use $M_\text{rep}$ to denote the privacy mechanism for token representation privatization. 
This process is referred to as \textit{perturbation} or \textit{noise injection}. 
To sample $N$ from the noise distribution, consider $N \in \mathbb{R}^{n}$ as a pair $(r, p)$, where $r$ is the distance from the origin and $p$ is a point in $\mathbb{B}^n$ (the unit hypersphere in $\mathbb{R}^{n}$). Then we  sample $N \in \mathbb{R}^{n}$ by computing $N=rp$, where $r$ is sampled from Gamma distribution $\Gamma(n, \frac{1}{\eta})$ and $p$ is sampled uniformly over $\mathbb{B}^n$~\cite{Wu2017BoltonDP,Fernandes2019GeneralisedDP}.

Although we describe $M_\text{rep}$ from the token's perspective, the same procedure also applies to sequence representations (e.g., the \texttt{[CLS]} representation produced by BERT for a sentence).

\subsubsection{\textbf{Text-to-Text Privatization}} \label{subsubsec:txt_privatization}
In addition to token representation privatization, we also study text-to-text privatization. We consider a \textit{token-to-token} case where each plain input token is transformed to a privatized output token. 
More formally, both the input and output of the randomized mechanism $M$ become a token $x,y\in\mathcal{V}$, where $\mathcal{V}$ is a vocabulary set. 
The privacy proof in \citet{Feyisetan2020PrivacyPreservingTA} show that $d\chi$-privacy can be achieved in this case by adding a post-processing step for $M_\text{rep}$ (Eq.~\ref{eqn:dx-N}) to map the output of $M_\text{rep}$ to another token via nearest neighbor search. More specifically, we first embed the input token $x$ using an embedding model $\phi: \mathcal{V}\rightarrow \mathbb{R}^n$. We then pass $\phi(x)$ to $M_\text{rep}$ to obtain the privatized token representation $M_\text{rep}(\phi(x))$. Lastly, we return the token that is closest to $M_\text{rep}(\phi(x))$ in the embedding space as the output,
\begin{equation}\label{eqn:dx-text}
\footnotesize
M_\text{txt}(x) = \mathop{\arg\min}_{w\in \mathcal{V}} {||M_\text{rep}(\phi(x))-\phi(w)||},
\end{equation}
where $M_\text{txt}$ denotes the privacy-preserving mechanism for text-to-text privatization. $M_\text{rep}$ and $M_\text{txt}$ offer equivalent privacy protection~\cite{Feyisetan2020PrivacyPreservingTA} since the post-processing strategy does not affect privacy guarantees. 
This process of nearest neighbor search is usually fast and scalable \cite{Feyisetan2020PrivacyPreservingTA} since the vocabulary can often fit in-memory and the operation can be optimized with ML accelerators.

Token representation privatization and text-to-text privatization each have their own merits. The former produces perturbed embeddings that could aid the neural models in exploiting the underlying semantics while the latter generates perturbed tokens that are easy to interpret for humans.
Also, text-to-text privatization is more compatible with existing pipelines of text processing~\cite{Feyisetan2020PrivacyPreservingTA}.

\subsubsection{\textbf{Sequence Input}} \label{subsubsec:seq-input}
Our discussion above considers the input $x$ as a single token (or representation for a single token). When the input becomes a sequence of tokens, $x=(x_i)_{1}^{\ell}$, we apply $M_\text{rep}$ or $M_\text{txt}$ on each token or token representation to privatize the sequence $x$ to $y=(y_i)_{1}^{\ell}$. In this case, the mechanism still satisfies $d\chi$-privacy, but for the distance function $d(x,x')=\sum_1^{\ell}||\phi(x_i)-\phi(x_i')||$. Further details and proofs can be found in  \citet{Feyisetan2020PrivacyPreservingTA}.

\subsection{Privacy-Constrained Fine-Tuning}
\label{subsec:approach-fine-tuning}
Due to the Local Privacy constraints, we do not access users' raw input text. However, we assume we have access to ground-truth labels of the NLU task since these labels can often be inferred from user behaviors, such as clicks~\cite{Zhang2019GenericIR} and other implicit feedback.

A neural NLU model typically consists of an embedding layer, an encoder, and task-specific layers (Fig.~\ref{fig:fine-tuning}). The embedding layer converts the text input to a sequence of token embeddings, which will then go through the encoder to produce a sequence representation. Finally, task-specific layers make predictions based on the sequence representation. 
We split the NLU model to user side and service-provider side 
to comply with Local Privacy constraints and discuss three privacy-constrained training/fine-tuning methods:

\begin{itemize}[leftmargin=1em]
    \item \textbf{Null privacy} (Fig.~\ref{fig:fine-tuning}.a). We do not apply any privacy constraints and thus cannot provide any privacy protections. This is to provide an upper bound for model utility.
 
    \item \textbf{Sequence representation privatization} (Fig.~\ref{fig:fine-tuning}.b). 
    The embedding layer and the encoder are deployed user-side. The user perturbs the sequence representation locally. 
    
    \item \textbf{Token representation privatization} (Fig.~\ref{fig:fine-tuning}.c). Only the token embedding layer is deployed at the user side. The user conducts tokenization and embedding table look-up locally to map the input text to token embeddings. They then privatize the token embeddings (Sec.~\ref{subsubsec:rep_privatization}) and send them to the service provider. Then service providers assemble the input sequence to the encoder by adding other necessary embeddings (e.g., positional embeddings) and injecting special tokens (e.g., \texttt{[CLS]}).
    
    \item \textbf{Text-to-text privatization} (Fig.~\ref{fig:fine-tuning}.d). The users conduct text-to-text privatization (Sec.~\ref{subsubsec:txt_privatization}) locally and send the privatized text to the service provider.
    Thus, service providers have a complete NLU model stack to process the privatized text.
\end{itemize}

Our foremost requirement is that the service provider only works with privatized input at both training and inference time, without any access to the raw user data. Thus, the service provider is not able to update the model parameters of user-side components. 
In contrast, the sequence representation privatization approach in
\citet{Lyu2020DifferentiallyPR} 
requires the service provider to access the raw user data during training, which violates the Local Privacy requirement at the training time. This fundamental difference makes our results incomparable with theirs.
Our pilot experiments indicate that sequence representation privatization yields undesirable utility (Sec.~\ref{subsec:results-fine-tuning}) when we make the user-side encoder untrainable to comply with the Local Privacy requirement. 
Thus, we focus on the other two privacy-constrained fine-tuning methods in this paper. 

In addition, we investigate the performance of two encoders, BERT and BiLSTM, to inspect the impact of different encoders under Local Privacy constraints. We mainly experiment with BERT since it is currently one of the most widely-used pretrained LMs. We also consider BiLSTM as a baseline encoder, which was used by \citet{Feyisetan2020PrivacyPreservingTA}. We use the same wordpiece embeddings for both encoders for fair comparisons.

\subsection{Privacy-Adaptive BERT Pretraining}
\label{subsec:approach-pretraining}
Inspired by the pretrained nature of BERT, we further propose privacy-adaptive pretraining methods to leverage a massive amount of unstructured texts that are publicly available. We also enjoy the flexibility of having access to the raw input in this case. These advantages of pretraining could make BERT more robust in handling privatized text or text representations. The pretrained model can also be used for different downstream tasks.

We initialize the model with the original BERT checkpoint and conduct further pretraining with the Next Sentence Prediction (NSP) loss~\cite{bert} and several variants of the Masked LM (MLM) loss we design. To simulate the scenario in privacy-constrained fine-tuning, we now assume the role of the users to produce large-scale privatized input for pretraining. We use the BooksCorpus~\cite{Zhu2015AligningBA} and the English Wikipedia data following BERT~\cite{bert}.

As explained in Sec~\ref{subsec:approach-fine-tuning}, user-side components cannot be updated by the service provider during privacy-constrained fine-tuning to comply with Local Privacy constraints. In the pretraining stage, we have the option to update user-side
components since we use public dataset for pretraining.
However, our pilot experiments indicate that, if user-side components are trainable, these components tend to generate representations that can be immune to perturbation.
For example, the updated token embedding layer tends to produce token embeddings that are less prone to be perturbed to a different token with the same $\eta$. Although this is good from the perspective of utility, this tendency severely affects the level of privacy protection demonstrated in Sec.~\ref{subsec:plausible_deniability_statistics} and \ref{subsec:token-embedding-inversion}. This privacy-utility trade-off motivates our decision that we must stop the gradient from being back-propagated to user-side components during pretraining so that we can maintain the same level of privacy protection while working on improving model utility. This measure also forces the model to adapt to perturbation with the model components on the service-provider side, instead of that on the user side. 

Since we have access to the raw input data during pretraining, we can take advantage of the MLM objective to train the BERT encoder to adapt to the perturbation process more effectively. We propose different privacy-adaptive pretraining methods based on different prediction targets of MLM as described below. 

\begin{itemize}[leftmargin=1em]
    \item \textbf{Vanilla MLM}: predicting the perturbed masked tokens. The most straightforward idea is to pretrain BERT on fully privatized corpora. This privatized LM could be more effective and robust in handling privatized content than the original LM. Since the privatization can be done on-the-fly during pretraining, the LM pretraining process benefits from seeing a diverse collection of perturbed text input. Formally, the vanilla MLM loss for a single masked position is defined as follows:
    \begin{equation}\label{eqn:vanilla-mlm}
    \footnotesize
    \begin{aligned}
    L_{\text{MLM}}^{\text{Vanilla}} = - \sum_{w^* \in \mathcal{V}} \mathds{1}\{w^*=\hat{w}\} &\log{\frac{\exp{\text{logit}(w^*)}}{\sum_{w' \in \mathcal{V}} \exp{\text{logit}(w')}}}
    \end{aligned}
    \end{equation}
    where $w^*$ is a candidate prediction of the privatized token and $\text{logit}(w^*)$ is the logit for making such a prediction. $\hat{w}$ is the true privatized token. $\mathds{1}\{\cdot\}$ is an indicator function.
    
    \item \textbf{Probability MLM} (Prob MLM): predicting a set of perturbed tokens for each masked position. d$\chi$-privacy guarantees that, for any finite $\eta$, the distribution of the perturbed tokens has a full support on the whole vocabulary~\cite{Feyisetan2020PrivacyPreservingTA}. In other words, every token in the vocabulary has a non-zero probability being selected as the perturbed token for a given input token. Meanwhile, the distribution of the perturbed tokens from d$\chi$-privacy remembers the semantics of the input token.
    Therefore, we perturb each masked token multiple times to obtain a set of perturbed tokens. The MLM losses coming from the perturbed tokens are weighted by their empirical frequencies. These empirical distributions of perturbed tokens could be beneficial for the LM to understand the injected noise, and thus, adapt to privatized content in a more efficient manner. Formally,
    \begin{equation}\label{eqn:prob-mlm}
    \footnotesize
    \begin{aligned}
    L_{\text{MLM}}^{\text{Prob}} = - \sum_{w^* \in \mathcal{V}} \frac{\text{count}(w^*, \hat{W})}{|\hat{W}|} &\log{\frac{\exp{\text{logit}(w^*)}}{\sum_{w' \in \mathcal{V}} \exp{\text{logit}(w')}}} 
    \end{aligned}
    \end{equation}
    where $\hat{W} = \{\hat{w}_i\}$ is a set of valid privatized tokens for this masked position and $|\hat{W}|$ denotes its size. We use $\text{count}(w^*, \hat{W})$ to denote the number of occurrence of a candidate prediction $w^*$ in $\hat{W}$.
    
    \item \textbf{Denoising MLM}: predicting the original tokens. Another idea is to let the model predict the original masked tokens so that the LM learns to recover the original semantics of the masked token given privatized context. Formally,
    \begin{equation}\label{eqn:denoising-mlm}
    \footnotesize
    \begin{aligned}
    L_{\text{MLM}}^{\text{Denoising}} = - \sum_{w^* \in \mathcal{V}} \mathds{1}\{w^*=w\} &\log{\frac{\exp{\text{logit}(w^*)}}{\sum_{w' \in \mathcal{V}} \exp{\text{logit}(w')}}}
    \end{aligned}
    \end{equation}
    where $w$ is the true original token for this masked position.
\end{itemize}

After further pretraining with our privacy-adaptive approaches, we fine-tune the BERT model with privacy-constrained training on task datasets (Sec.~\ref{subsec:approach-fine-tuning}) to evaluate the final model performance.
\section{Datasets}
\label{sec:datasets}
As listed below, we use two datasets from the GLUE benchmark~\cite{Wang2018GLUEAM} for our privacy and utility experiments. This benchmark was also used to evaluate BERT~\cite{bert} and other popular NLU models, including XLNet~\cite{Yang2019XLNetGA}, and RoBERTa~\cite{Liu2019RoBERTaAR}. 
\begin{itemize}[leftmargin=1em]
\item \textbf{Stanford Sentiment Treebank (SST)}\footnote{\url{https://nlp.stanford.edu/sentiment/index.html}}~\cite{sst} is a single sentence classification task. The goal is to predict a sentiment label (positive or negative) for a sentence of movie review. This dataset contains 67k training sentences and 872 validation sentences.

\item \textbf{Quora Question Pairs (QQP)}\footnote{\url{https://www.quora.com/q/quoradata/First-Quora-Dataset-Release-Question-Pairs}} is a sentence-pair classification task. The goal is to determine whether a pair of questions are paraphrases or not. This dataset is larger than SST and has 363k sentence pairs for training and 40k sentence pairs for validation.
\end{itemize} 

Our adoption of these public datasets contributes to the reproducibility of our paper. Our selection also covers both the single sentence classification task and sentence pair classification task. 
Since the GLUE test sets are hidden, we train our models and baselines on the training set and report the results on the validation set. We use accuracy as the metric for utility experiments in Sec.~\ref{sec:utility-exp}.

\section{Privacy Experiments}
\label{sec:privacy-experiments}
We present a series of privacy experiments to demonstrate the level of privacy protection provided by the privacy mechanism on BERT, which has not been studied before to the best of our knowledge. 

\subsection{Geometry of the BERT Embedding Space}
\label{subsec:geometry}

As mentioned in Sec.~\ref{subsec:preliminaries}, the level of privacy protection from $d\chi$-privacy also depends on the distance function $d$, which is defined as the Euclidean distance of tokens in the BERT embedding space for our token privatization mechanisms (Sec.~\ref{subsubsec:rep_privatization} and Sec.~\ref{subsubsec:txt_privatization}). Thus, to understand the privacy protection and the noisy injection of these mechanisms, we analyze the geometry properties of the BERT embedding space in this section.

\begin{figure}[t]
    \centering
    \includegraphics[width=0.4\textwidth]{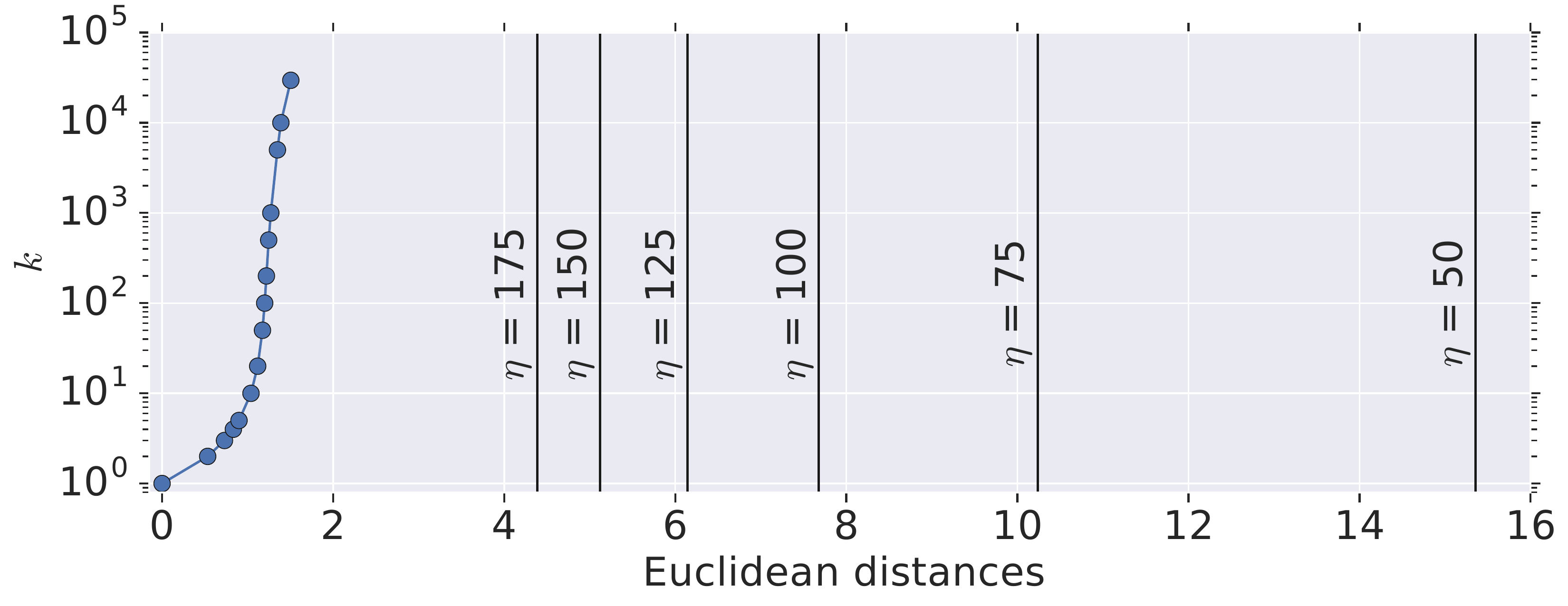}
    \vspace{-0.5cm}
    \caption{Geometry of the BERT embedding space. 
    Dots are average Euclidean distances of an original embedding and its $k$ nearest neighbors. Vertical lines are average Euclidean distances between original and perturbed embeddings. 
    }
    \label{fig:geometry}
\end{figure}

We first compute the Euclidean distance between an original token embedding and its privatized/perturbed token embedding, i.e., $|| x-M_\text{rep}(x)||=||N||$ from Eq.~\ref{eqn:dx-N}. We average these distances for all tokens in the BERT vocabulary, and compute this average distance for a set of strategically chosen $\eta$ values. The results are shown as the black vertical lines in Fig.~\ref{fig:geometry} labeled by the corresponding $\eta$. 
As expected, we observe that the average Euclidean distance corresponds to the mean of the Gamma distribution (Sec.~\ref{subsubsec:rep_privatization}), i.e., $\frac{n}{\eta}$, where $n=768$ is the dimension size of BERT token embeddings. 
As $\eta$ becomes smaller, the average distance grows progressively larger, implying increasingly larger noise.

Next, we compute the Euclidean distances among the original token embeddings for a calibrated comparison. Specifically, we compute the distances between each token and its $k$-th nearest neighbor, for all tokens in the BERT vocabulary $\mathcal{V}$. We then compute the average $k$-th nearest neighbor distance, for each $k$ in [1, 2, 3, 4, 5, 10, 20, 50, 100, 200, 500, 1000, 5000, 10000, $|\mathcal{V}|$]. The results are presented as blue dots in Fig.~\ref{fig:geometry}. As $k$ becomes larger, the average Euclidean distance also grows larger, as expected.

Finally, we compare the distances computed in the previous steps. Distances among the original token embeddings are smaller than 2 on average while even the smallest noise we used in experiments ($\eta = 175$) perturbs tokens to positions that are as twice as far on average. This gives us a sense of how large these perturbations are. Sec.~\ref{subsec:token-embedding-inversion} justifies using such large noise by showing nearest neighbor search in the embedding space (Sec.~\ref{subsubsec:txt_privatization}) can still map the perturbed token embedding to its original token. Our choices of $\eta$ are greater than the typically used $\epsilon$ in LDP. Given the relatively high dimensionality of BERT embeddings (768) and the large noise we require, we adopt higher $\eta$ to achieve the level of privacy protection shown in the next sections. To bring down the level of $\eta$ to values resembling $\varepsilon$ in LDP, one may reduce the dimensionality of the BERT embeddings by random projection~\cite{Wu2017BoltonDP}, which will be investigated in our future work.

\subsection{Examples of Perturbed Text}
\label{subsec:example-of-perturbed-text}
We present examples of text-to-text privatization (Sec.~\ref{subsubsec:txt_privatization}) in Tab.~\ref{tab:perturbed-text} with different choices of $\eta$. As expected, we observe that as $\eta$ becomes larger, the noise becomes smaller and more tokens remain unmodified.
Next, we focus on $\eta=100$ (as an example), and perturb each token in the previous sentence 1,000 times. We sort the top-10 perturbed tokens for each original token by their empirical frequencies. The results are presented in Tab.~\ref{tab:perturbed-text-support}.
We observe that a token has a certain probability to be preserved. For example, ``emotions'' has 14\% of the chances to remain unmodified under this specific choice of $\eta$. A token can also be perturbed into a diverse set of other tokens. For instance, ``emotions'' is perturbed to 825 distinct words with different empirical frequencies. Some of the top words are ``emotion'', ``emotionally'', ``hormones'', ``emotional'', and ``moods'', which share similar semantics with the original word ``emotions''. Some perturbed tokens might not have obvious links to the original token due to the randomness introduced by the privacy-preserving mechanism. Intuitively, good privacy protection implies a token has a relatively small chance of being preserved, and can be modified to a relatively diverse set of perturbed tokens. We quantify these measurements in the next section.

\setlength{\tabcolsep}{2pt}
\begin{table}[t]
\caption{Examples of perturbed sentences with different choices of $\eta$. ``Orig'' denotes the original input sentence. The red color denotes tokens that are modified.}
\label{tab:perturbed-text}
\vspace{-0.4cm}
\footnotesize
\begin{tabular}{@{}c|lllllllll@{}}
\toprule
$\eta$ & \multicolumn{9}{c}{Sequence}                                                                                                                                                                                                                                                                                   \\ \midrule
Orig    & the                            & emotions                        & are                                & raw                              & and                         & will                            & strike                           & a                             & nerve                            \\ \midrule
50      & {\color[HTML]{FF0000} \#\#ori} & {\color[HTML]{FF0000} backward} & {\color[HTML]{FF0000} og}          & {\color[HTML]{FF0000} wanda}     & {\color[HTML]{FF0000} big}  & {\color[HTML]{FF0000} disposal} & {\color[HTML]{FF0000} \#\#pose}  & {\color[HTML]{FF0000} lou}    & {\color[HTML]{FF0000} \#\#bular} \\
75      & {\color[HTML]{FF0000} 410}     & {\color[HTML]{FF0000} truth}    & {\color[HTML]{FF0000} go} & {\color[HTML]{FF0000} mole}      & {\color[HTML]{FF0000} gone} & will                            & strike                           & {\color[HTML]{FF0000} y}      & {\color[HTML]{FF0000} gifford}   \\
100     & {\color[HTML]{FF0000} fine}    & {\color[HTML]{FF0000} abused}   & are                                & {\color[HTML]{FF0000} primitive} & {\color[HTML]{FF0000} it}   & will                            & {\color[HTML]{FF0000} slaughter} & {\color[HTML]{FF0000} us}     & nerve                            \\
125     & {\color[HTML]{FF0000} measure} & emotions                        & :                                  & {\color[HTML]{FF0000} shield}    & and                         & {\color[HTML]{FF0000} relation} & strike                           & {\color[HTML]{FF0000} nearly} & nerve                            \\
150     & the                            & {\color[HTML]{FF0000} caleb}    & are                                & {\color[HTML]{FF0000} kill}      & and                         & will                            & strike                           & {\color[HTML]{FF0000} circle} & nerve                            \\
175     & the                            & emotions                        & are                                & raw                              & and                         & will                            & strike                           & a                             & nerve                            \\ \bottomrule
\end{tabular}
\end{table}
\setlength{\tabcolsep}{1.5pt}
\begin{table}[t]
\caption{Examples of perturbed tokens ($\eta=100$, sampled independently and sorted by empirical frequencies).}
\label{tab:perturbed-text-support}
\vspace{-0.3cm}
\footnotesize
\begin{tabular}{@{}l|l|l|l|l|l|l|l|l|l@{}}
\toprule
                        & the  & emotions      & are   & raw        & and & will   & strike  & a   & nerve          \\ \midrule
\multirow{7}{*}{\rotatebox{90}{Perturbed tokens}} & the  & emotions      & are   & raw        & and & will   & strike  & a   & nerve          \\
                           & a    & emotion       & were  & smackdown  & or  & would  & strikes & the & rebels         \\
                           & its  & emotionally   & is    & matt       & but & can    & attack  & an  & reason         \\
                           & and  & hormones      & being & \#\#awa    & ,   & may    & drop    & his & cells          \\
                           & his  & organizations & re    & unused     & -   & better & \#\#gen & its & spirits        \\
                           & her  & emotional     & have  & division   & as  & must   & aim     & her & bothering      \\
                           & some & moods         & of    & protection & "   & self   & stroke  & one & communications \\ \bottomrule
\end{tabular}
\end{table}
\subsection{Plausible Deniability Statistics}
\label{subsec:plausible_deniability_statistics}

We follow \citet{Feyisetan2020PrivacyPreservingTA} to use two statistics to characterize the ability of an adversary to recover the original input text when observing the perturbed text or text representations. This ability is referred to as plausible deniability~\cite{Bindschaedler2017PlausibleDF} and it varies under different settings of $\eta$. The formal definitions of the statistics can be found in \citet{Feyisetan2020PrivacyPreservingTA}. We provide intuitive explanations as follows:
\begin{itemize}[leftmargin=1em]
    \item $N_w$: the probability of an input token $w$ not modified by $M_\text{txt}(w)$.
    
    \item $S_w$: the effective support of the output distribution of perturbation $M_\text{txt}(w)$ on the entire vocabulary for an input token $w$.  
\end{itemize}

We run simulations to estimate the plausible deniability statistics. For each choice of privacy parameter $\eta$, we perturb each regular token in the vocabulary for 1,000 times. Regular tokens refer to the tokens other than \texttt{[PAD]}, \texttt{[CLS]}, \texttt{[SEP]}, \texttt{[MASK]}, \texttt{[UNK]}, or \texttt{[unused...]}. We estimate $N_w$ as the number of output tokens that are identical to the input token, and $S_w$ as the number of unique output tokens. Intuitively, good privacy guarantees should be characterized by relatively small $N_w$ and relatively large $S_w$.

\begin{figure*}[t]
    \centering
    \includegraphics[width=0.8\textwidth]{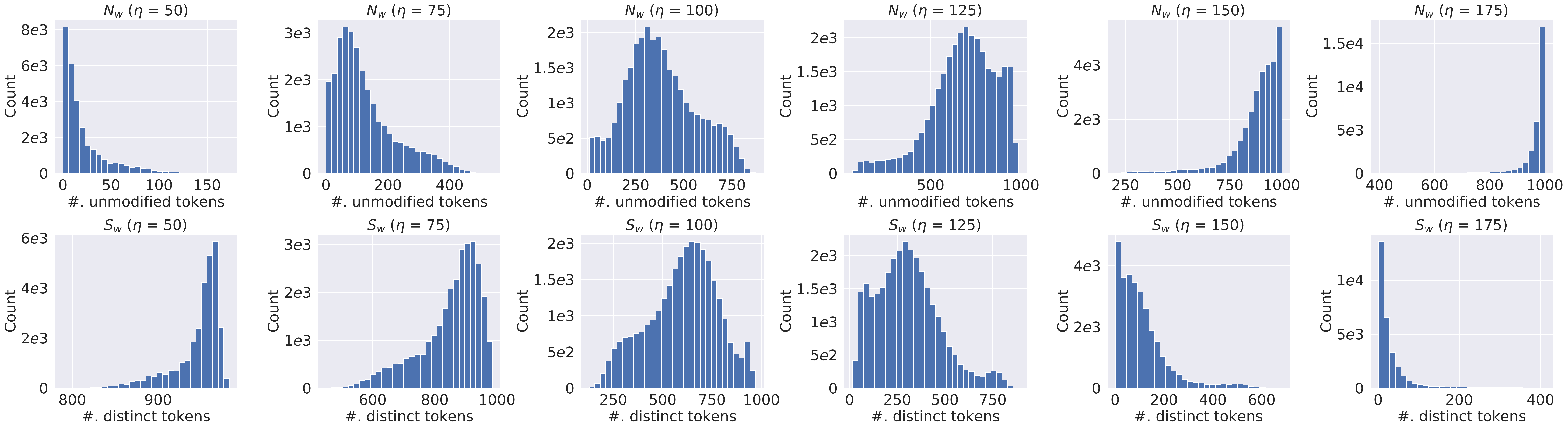}
    \vspace{-0.2cm}
    \caption{Plausible deniability statistics. Each token in the vocabulary is perturbed 1,000 times. $N_w$ refers to the number of output tokens that are identical to the input token, and $S_w$ refers to the number of unique output tokens.}
    \label{fig:plausible_deniability_statistics}
\end{figure*}

We present the estimated plausible deniability statistics of the BERT vocabulary in Fig.~\ref{fig:plausible_deniability_statistics}. 
For example, when $\eta=75$, the $N_w$ figure shows no token is ever returned more than 500 times in the worst case and the $S_w$ figure shows no token produces fewer than 500 distinct new tokens.
As $\eta$ becomes greater (the noise becomes smaller), a growing number of tokens tend to have larger $N_w$, indicating that the tokens tend to have greater probability of remaining unmodified. Meanwhile, more and more tokens tend to produce only a small amount of distinct perturbed outputs ($S_w$), suggesting a limited support on the vocabulary.

These statistic figures serve as a visual guidance for selecting $\eta$ for different applications that require different privacy guarantees. \citet{Feyisetan2020PrivacyPreservingTA} suggest selecting $\eta$ based on the desired worst case guarantees. 
For example, we should select $\eta$ as 75 for an application that requires all tokens to be modified for at least 500 times out of 1,000 perturbations.
In addition to the worst case guarantees, we show another practical approach to conduct $\eta$ selection based on the average case guarantees in the next section.

\subsection{Token Embedding Inversion}
\label{subsec:token-embedding-inversion}

The plausible deniability statistics characterize the privacy guarantees from the perspective of the vocabulary. In this section, we show another interpretation of privacy protection level by measuring the amount of tokens that are modified by perturbation in actual datasets following \citet{Song2020InformationLI}. This analysis accounts for the token frequencies in real datasets and could be more practical in guiding the selection of the privacy parameter $\eta$.

We define an adversarial task, token embedding inversion, as recovering the original tokens based on perturbed token embeddings. We leverage nearest neighbor search for this task. Given a perturbed token embedding, we find the nearest neighbor of this embedding in the embedding space as the predicted original token. The performance is measured by accuracy. Although this attack is not comprehensive, it's performance can be considered as an automatic metric to reveal the privacy-utility trade-off and to guide the selection of the privacy parameter $\eta$. Besides, this approach is shown to be highly effective on our low-level lexical input (tokens). We will leave more advanced ML attacks to our future work.

The results of token embedding inversion on the validation data of SST and QQP are presented in Tab.~\ref{tab:token-embedding-inverstion}. We demonstrate that privacy protection levels are consistent across datasets, indicating that the privacy-preserving mechanism could be agnostic to specific tasks and datasets. As $\eta$ becomes larger and the noise becomes smaller, a growing amount of tokens in actual datasets can be correctly recovered by nearest neighbor search. For example, at $\eta = 100$, only 34\% of tokens in the datasets can be recovered, indicating that this choice of $\eta$ provides moderately strong privacy protection. 

\setlength{\tabcolsep}{4.5pt}
\begin{table}[t]
\caption{Accuracy of token embedding inversion.}
\label{tab:token-embedding-inverstion}
\vspace{-0.3cm}
\footnotesize
\begin{tabular}{@{}c|cccccc@{}}
\toprule
$\eta$ & 50     & 75     & 100    & 125    & 150    & 175    \\ \midrule
SST     & 0.0154 & 0.1084 & 0.3402 & 0.6354 & 0.8500 & 0.9511 \\
QQP     & 0.0165 & 0.1066 & 0.3420 & 0.6462 & 0.8620 & 0.9570 \\ \bottomrule
\end{tabular}
\end{table}

We also observe that the inversion scores roughly correspond to the peak values of $N_w$ in plausible deniability statistics (Fig.~\ref{fig:plausible_deniability_statistics}). This observation advocates for using average case guarantees for $\eta$ selection. $\eta$ selection with worst case guarantees always favors stricter privacy choices since this approach is contingent on the tokens that are least prone to be modified. 
$\eta$ selection based on average case guarantees is an alternative approach that accounts for the term frequencies in actual corpora. 
Our recommendation is that, given the different linguistic properties in different languages, domains, applications, and data, the decision on the choice of $\eta$ and the approach to select an $\eta$ should be made on a case-by-case basis.

\section{Utility Experiments}
\label{sec:utility-exp}

\subsection{\textbf{Implementation Details}}
\label{subsec:implementation}

\textbf{Privacy-Constrained Fine-Tuning.}
We freeze parameters for a model component to simulate the user-side deployment for this component. E.g., in token representation privatization, the embedding layer is frozen since it is deployed at the user side (Fig.~\ref{fig:fine-tuning}).
For both BERT and BiLSTM, we set the maximum sequence length to 128, the training batch size to 32, and the number of training epochs to 3. On SST, we use the learning rate of 2e-5 to fine-tune BERT and 1e-3 to train BiLSTM. On QQP, we use the learning rate of 4e-5 to fine-tune BERT and 1e-3 to train BiLSTM. The warm up portion of the learning rate is 10\% of the total steps. We use the same wordpiece embeddings for both encoders for fair comparisons. 

\textbf{Privacy-Adaptive BERT Pretraining.}
We set the maximum sequence length to 128, the training batch size to 256, the learning rate to 2e-5, the maximum number of predictions per sequence to 20, the mask rate to 0.15, the maximum training steps to 1,000,000, and warm up steps to 10,000. We save checkpoints every 200,000 steps and fine-tune the checkpoint with privacy-constrained training to select the best checkpoint. The choices of $\eta$ for pretraining and fine-tuning are identical in our experiments. For Prob MLM, we set the number of perturbations per masked position to 10.

\subsection{Results \nolinebreak of \nolinebreak Privacy-Constrained \nolinebreak Fine-Tuning}
\label{subsec:results-fine-tuning}
Our experiments on sequence representation privatization indicates even the smallest perturbation we used ($\eta$=175) causes 30\% absolute performance decrease compared with Null Privacy. Given this observation, we focus on experiments for token-level privatization.\footnote{We should note that the privacy semantics of $\eta$ for sequence-level privatization and token-level privatization are different. We leave this to our future work.}

The utility results for token-level privatization are reported in Tab.~\ref{tab:fine-tuning-results-sst} for SST and Tab.~\ref{tab:fine-tuning-results-qqp} for QQP. We use the same wordpiece embeddings for both BERT and BiLSTM encoders for fair comparisons. The privacy parameters $\eta$ for token-level privatization, including those with privacy-adaptive pretraining, are directly comparable since they use equivalent inputs and distance functions.
    
We first compare the two token-level privatization methods (column 1 vs. 2, column 3 vs. 4 in Tab.~\ref{tab:fine-tuning-results-sst} and \ref{tab:fine-tuning-results-qqp}). We see text-to-text privatization produces considerable performance improvement compared to token representation privatization. This improvement has statistical significance and is observed for both encoders on a wide range of $\eta$. The performance gain is larger when noise is smaller. This verifies our finding in Sec.~\ref{subsec:token-embedding-inversion} that nearest neighbor search is effective in mapping perturbed embeddings to its original token (although this does not affect the privacy guarantees~\cite{Feyisetan2020PrivacyPreservingTA}). 
\citet{Feyisetan2020PrivacyPreservingTA} take advantage of this technique to make the privatization mechanism compatible with existing pipelines of text processing, while we reveal the important role it plays to dramatically improve the performance of NLU models in privacy-constrained fine-tuning.

We then look at the comparisons of the two encoders. BERT performs better than BiLSTM if we do not apply privacy constraints, as expected. We observe that BERT suffers from a larger performance degradation than BiLSTM with token representation privatization. This indicates that BERT is less robust than BiLSTM in handling perturbation on representations. This unexpected result can be explained by the mismatch of representations observed by BERT during pretraining (on plain corpora) and fine-tuning (on privatized corpora). Also, complex models are more likely to suffer from inherently high variance. In the next section, we show our privacy-adaptive pretraining methods can improve BERT's performance and help it outperform BiLSTM under the same privacy guarantee. 
On the other hand, BERT performs better with
text-to-text privatization, compared with BiLSTM, in general. Thanks to the advantages brought by the LM pretraining and attention mechanisms, BERT handles the perturbation better than BiLSTM with statistical significance for a wide range of $\eta$ values. This shows BERT is more robust than BiLSTM in handling text-to-text privatization.

Finally, we briefly look at the comparison of our approach with Null Privacy. Perturbation on token representations results in severe degradation in model performance compared with Null Privacy. Even though the encoder is fine-tuned in token representation privatization, it is not able to adapt to the perturbed token embeddings. This is within our expectation since we observe in Sec.~\ref{subsec:geometry} that the noise applied to embeddings causes a significant shift of the original embedding in the embedding space. Text-to-text privatization, on the other hand, shows less degradation. 

\begin{table}[t]
\caption{Accuracy of privacy-constrained fine-tuning on SST. Scores in the header are obtained with Null Privacy. Boldface denotes better results with the same encoder. Underscores denote better results between different encoders with the same privacy constraint. $\blacktriangle i$ denotes the improvement with respect to column $i$ has statistically significance with $p < 0.05$ tested by the Student’s paired t-test.
}
\label{tab:fine-tuning-results-sst}
\vspace{-0.3cm}
\footnotesize
\begin{tabular}{@{}l|ll|ll@{}}
\toprule
\multirow{2}{*}{$\eta$} & \multicolumn{2}{c|}{BERT (0.9289)}            & \multicolumn{2}{c}{BiLSTM (0.8406)}                                \\ \cmidrule(l){2-5} 
                            & 1. Token Rep             & 2. Text-to-Text                  & 3. Token Rep                   & 4. Text-to-Text          \\ \midrule
50                          & {\ul \textbf{0.5126}} & 0.4920                        & \textbf{0.5092}       & {\ul \textbf{0.5092}} \\
75                          & 0.5310                & {\ul \textbf{0.5906}}$^{\blacktriangle 1,4}$ & {\ul \textbf{0.5447}} & 0.5356                \\
100                         & 0.5298                & {\ul \textbf{0.7030}}$^{\blacktriangle 1,4}$ & {\ul 0.5608}          & \textbf{0.6697}$^{\blacktriangle 3}$       \\
125                         & {\ul 0.5608}          & {\ul \textbf{0.8360}}$^{\blacktriangle 1,4}$ & {\ul 0.5608}          & \textbf{0.7420}$^{\blacktriangle 3}$       \\
150                         & 0.5665                & {\ul \textbf{0.8968}}$^{\blacktriangle 1,4}$ & {\ul 0.5917}          & \textbf{0.8119}$^{\blacktriangle 3}$       \\
175                         & 0.5975                & {\ul \textbf{0.9151}}$^{\blacktriangle 1,4}$ & {\ul 0.6216}          & \textbf{0.8211}$^{\blacktriangle 3}$       \\ \bottomrule
\end{tabular}
\end{table}

\begin{table}[t]
\caption{Accuracy of privacy-constrained fine-tuning on QQP. Refer to Tab.~\ref{tab:fine-tuning-results-sst} to interpret the notations.}
\label{tab:fine-tuning-results-qqp}
\vspace{-0.3cm}
\footnotesize
\begin{tabular}{@{}l|ll|ll@{}}
\toprule
\multirow{2}{*}{$\eta$} & \multicolumn{2}{c|}{BERT (0.9106)}          & \multicolumn{2}{c}{BiLSTM (0.8261)}  \\ \cmidrule(l){2-5} 
                         & 1. Token Rep           & 2. Text-to-Text                & 3. Token Rep        & 4. Text-to-Text             \\ \midrule
50                       & \textbf{0.6370}$\blacktriangle^2$ & 0.6318                      & {\ul 0.6409}$^{\blacktriangle 1}$ & {\ul \textbf{0.6423}}$^{\blacktriangle 2}$  \\
75                       & {\ul 0.6318}        & \textbf{0.6485}$^{\blacktriangle 1}$         & {\ul 0.6318}     & {\ul \textbf{0.6631}}$^{\blacktriangle 2,3}$  \\
100                      & {\ul 0.6318}        & {\ul \textbf{0.7238}}$^{\blacktriangle 1,4}$ & {\ul 0.6318}     & \textbf{0.7108}$^{\blacktriangle 3}$        \\
125                      & {\ul 0.6318}        & {\ul \textbf{0.8274}}$^{\blacktriangle 1,4}$ & {\ul 0.6318}     & \textbf{0.7534}$^{\blacktriangle 3}$        \\
150                      & 0.6447              & {\ul \textbf{0.8759}}$^{\blacktriangle 1,4}$ & {\ul 0.6811}$^{\blacktriangle 1}$ & \textbf{0.7854}$^{\blacktriangle 3}$        \\
175                      & 0.6354              & {\ul \textbf{0.8976}}$^{\blacktriangle 1,4}$ & {\ul 0.6987}$^{\blacktriangle 1}$ & \textbf{0.8066}$^{\blacktriangle 3}$        \\ \bottomrule
\end{tabular}
\end{table}

\subsection{Results of Privacy-Adaptive Pretraining}
\label{subsec:pretraining}
\begin{table*}[t]
\caption{Accuracy of privacy-adaptive pretraining with token representation privatization. Italic denotes better results than the original BERT. Underscores denote better results than the pretraining baseline of Vanilla MLM. Boldface denotes the best results. $\blacktriangle i$ denotes the gain with respect to column $i$ has statistically significance ($p < 0.05$ tested by the Student’s paired t-test).}
\label{tab:pretrain-token-rep}
\vspace{-0.4cm}
\footnotesize
\begin{tabular}{@{}l|llll|llll@{}}
\toprule
\multirow{2}{*}{$\eta$} & \multicolumn{4}{c|}{SST}                                                                     & \multicolumn{4}{c}{QQP}                                                                         \\ \cmidrule(l){2-9} 
                            & 1. Orig BERT & 2. Vanilla MLM                        & 3. Prob MLM                                            & 4. Denoising MLM                                           & 1. Orig BERT & 2. Vanilla MLM                               & 3. Prob MLM             & 4. Denoising MLM                  \\ \midrule
50                          & 0.5126      & \textit{0.5390}$^{\blacktriangle 1}$ & {\ul \textit{\textbf{0.5424}}}$^{\blacktriangle 1}$ & \textit{0.5356}                                         & 0.6370 & \textit{0.6434}$^{\blacktriangle 1,3}$          & 0.6364                                  & {\ul \textit{\textbf{0.6444}}}$^{\blacktriangle 1,3}$ \\
75                          & 0.5310      & \textit{0.5791}$^{\blacktriangle 1}$ & \textit{0.5757}$^{\blacktriangle 1}$                & {\ul \textit{\textbf{0.6089}}}$^{\blacktriangle 1}$     & 0.6318 & \textit{0.6645}$^{\blacktriangle 1}$          & {\ul \textit{0.6821}}$^{\blacktriangle 1,2}$ & {\ul \textit{\textbf{0.7119}}}$^{\blacktriangle 1,2,3}$ \\
100                         & 0.5298      & \textit{0.6709}$^{\blacktriangle 1}$ & {\ul \textit{0.6766}}$^{\blacktriangle 1}$          & {\ul \textit{\textbf{0.7041}}}$^{\blacktriangle 1,2}$   & 0.6318 & \textit{0.7668}$^{\blacktriangle 1}$          & {\ul \textit{0.7686}}$^{\blacktriangle 1}$ & {\ul \textit{\textbf{0.7788}}}$^{\blacktriangle 1,2,3}$ \\
125                         & 0.5608      & \textit{0.7706}$^{\blacktriangle 1}$ & {\ul \textit{0.7718}}$^{\blacktriangle 1}$          & {\ul \textit{\textbf{0.7810}}}$^{\blacktriangle 1}$     & 0.6318 & \textit{0.8200}$^{\blacktriangle 1}$          & {\ul \textit{0.8219}}$^{\blacktriangle 1}$ & {\ul \textit{\textbf{0.8249}}}$^{\blacktriangle 1,2,3}$ \\
150                         & 0.5665      & \textit{0.8188}$^{\blacktriangle 1}$ & \textit{0.8188}$^{\blacktriangle 1}$                & {\ul \textit{\textbf{0.8395}}}$^{\blacktriangle 1,2,3}$ & 0.6447 & \textit{0.8520}$^{\blacktriangle 1}$          & \textit{0.8520}$^{\blacktriangle 1}$       & {\ul \textit{\textbf{0.8523}}}$^{\blacktriangle 1}$ \\
175                         & 0.5975      & \textit{0.8658}$^{\blacktriangle 1}$ & {\ul \textit{\textbf{0.8693}}}$^{\blacktriangle 1}$ & {\ul \textit{\textbf{0.8693}}}$^{\blacktriangle 1}$     & 0.6354 & \textit{\textbf{0.8698}}$^{\blacktriangle 1}$ & \textit{0.8688}$^{\blacktriangle 1}$       & \textit{0.8691}$^{\blacktriangle 1}$                \\ \bottomrule
\end{tabular}
\end{table*}

\begin{table*}[t]
\caption{Accuracy of privacy-adaptive pretraining with token-to-token privatization. 
Refer to Tab.~\ref{tab:pretrain-token-rep} to interpret the notations.
}
\label{tab:pretrain-text-to-text}
\vspace{-0.4cm}
\footnotesize
\begin{tabular}{@{}l|llll|llll@{}}
\toprule
\multirow{2}{*}{$\eta$} & \multicolumn{4}{c|}{SST}                                                                                 & \multicolumn{4}{c}{QQP}                                                                                  \\ \cmidrule(l){2-9} 
                            & 1. Orig BERT & 2. Vanilla MLM              & 3. Prob MLM                       & 4. Denoising MLM                  & 1. Orig BERT & 2. Vanilla MLM              & 3. Prob MLM                       & 4. Denoising MLM                  \\ \midrule
50                          & 0.4920      & \textit{0.5218}          & {\ul \textit{\textbf{0.5310}}} & \textit{0.5092}                & 0.6318      & \textit{0.6375}$^{\blacktriangle 1}$          & {\ul \textit{0.6419}}$^{\blacktriangle 1,2}$        & {\ul \textit{\textbf{0.6469}}}$^{\blacktriangle 1,2,3}$ \\
75                          & 0.5906      & \textit{\textbf{0.5963}} & 0.5734                         & \textit{0.5906}                & 0.6485      & \textit{0.6643}$^{\blacktriangle 1}$          & \textit{0.6616}$^{\blacktriangle 1}$                & {\ul \textit{\textbf{0.6693}}}$^{\blacktriangle 1,2,3}$ \\
100                         & 0.7030      & \textit{0.7190}          & {\ul \textit{\textbf{0.7259}}} & {\ul \textit{0.7225}}          & 0.7238      & \textit{\textbf{0.7628}}$^{\blacktriangle 1}$ & \textit{0.7610}$^{\blacktriangle 1}$                & \textit{0.7603}$^{\blacktriangle 1}$                \\
125                         & 0.8360      & \textit{\textbf{0.8429}} & {\ul \textit{\textbf{0.8429}}} & \textit{0.8406}                & 0.8274      & \textit{0.8346}$^{\blacktriangle 1}$          & {\ul \textit{\textbf{0.8362}}}$^{\blacktriangle 1}$ & {\ul \textit{0.8351}}$^{\blacktriangle 1}$          \\
150                         & 0.8968      & \textit{0.9048}          & {\ul \textit{\textbf{0.9071}}} & \textit{0.9025}                & 0.8759      & \textit{0.8790}$^{\blacktriangle 1}$          & {\ul \textit{\textbf{0.8801}}}$^{\blacktriangle 1}$ & {\ul \textit{0.8791}}$^{\blacktriangle 1}$          \\
175                         & 0.9151      & \textit{0.9209}          & {\ul \textit{0.9209}}          & {\ul \textit{\textbf{0.9220}}} & 0.8976      & \textit{\textbf{0.8997}}                      & \textit{0.8994}                                     & \textit{0.8990}                \\ \bottomrule
\end{tabular}
\end{table*}

Privacy-adaptive pretraining results are shown in Tab.~\ref{tab:pretrain-token-rep} for token representation privatization and Tab.~\ref{tab:pretrain-text-to-text} for text-to-text privatization.

We first analyze the general effect of privacy-adaptive pretraining. For token representation privatization, consistent results on two datasets demonstrate that all three privacy-adaptive pretraining methods have significant performance improvement with statistical significance. This suggests that the BERT pretrained with privacy-adaptive methods are much more effective and robust than the original BERT model in handling privatized token representations. For text-to-text privatization, although it has inherently advantages due to nearest neighbor search, privacy-adaptive pretraining still manages to outperform the original BERT checkpoint on both datasets, and with statistical significance on QQP. These results demonstrate the effectiveness of our privacy-adaptive pretraining.

We then compare the performance of the three privacy-adaptive pretraining approaches. For token representation privatization, the Vanilla MLM is already highly effective, despite its simplicity. It only brings a marginal improvement if we enable the model to predict a set of perturbed tokens for each masked position with Prob MLM. This indicates that the augmentation effect in Vanilla MLM could be sufficient for the model to observe enough variations of the noise injection process. Predicting the original tokens with Denoising MLM shows visible gains compared with predicting perturbed tokens in Vanilla and Prob MLM. These improvements are particularly compelling and have statistical significance when we have moderately large noise ($\eta$ = 75, 100), where more than half of the tokens in the datasets are modified. These results indicate Denoising MLM has more practical values in real-world applications. 

For text-to-text privatization, Denoising MLM remains a competitive privacy-adaptive pretraining methods. In contrast to token representation privatization, predicting (a set of) perturbed tokens here shows marginal improvement over predicting the original token in some cases. Since text-to-text privatization sometimes lands on the original token as the perturbed token due to nearest neighbor search, Denoising MLM is less beneficial than the case with token representation privatization. In particular, privacy-adaptive pretraining with text-to-text privatization demonstrates considerable improvement when $\eta$ = 100, where the noise is sufficiently large and more than half of the tokens in the datasets are modified. This improvement expands to a wider range of $\eta$ values ($\eta$ = 50, 75, 100, 125) on QQP. Privacy-adaptive pretraining has a pronounced effect on QQP, probably because the sentence pair prediction task is more difficult, and thus, is more dependent on the adaptation of perturbation in our privacy-adaptive pretraining process. This demonstrates the value of our methods in practical applications.

Finally, we compare the performance of privacy-adaptive pretraining between the two different token-level privatization. It is worth noting that token representation privatization generally outperforms text-to-text privatization when noise is large ($\eta$ = 50, 75, and 100 in some cases). 
This exciting observation indicates that the noise adaptation effect resulting from privacy-adaptive pretraining has advantages over nearest neighbor search when noise is large.

To sum up, we recommend adopting Denoising MLM as the primary method for privacy-adaptive pretraining for both token-level privatization approaches. When a relatively strong level of privacy protection (e.g., $\eta < 100$) is required, token representation privatization should be adopted to preserve more utility. Otherwise, text-to-text privatization is preferred. In both cases, privacy-adaptive pretraining is essential to improve model performance.
\section{Conclusions and Future Work} 
\label{sec:conclusion}
In this work, we study how to improve NLU model performance on privatized text in the Local Privacy setting. 
We first take a deep analytical view to illustrate the privacy guarantees of $d\chi$ privacy. We then investigate the behavior of BERT when it meets privatized text input with privacy-constrained fine-tuning methods. 
We show BERT is less robust than BiLSTM in handling privatized token representations. We further show text-to-text privatization can often improve upon token representation privatization, revealing the important role played by nearest neighbor search to improve utility in privacy-constrained fine-tuning. 
More importantly, we propose privacy-adaptive LM pretraining methods and demonstrate that a BERT pretrained with our Denoising MLM objective is more robust in handling privatized content compared with the original BERT. 
For future work, we would like to study privatization on contextualized token representations and more advanced ML attacks. We will also look at reducing the dimensionality of the BERT embeddings by random projection~\cite{Wu2017BoltonDP} to bring down the level of $\eta$.

\begin{acks}
The authors would like to thank Borja Balle for the helpful discussions and constructive comments on this work.
\end{acks}

\newpage
\bibliographystyle{ACM-Reference-Format}
\balance 
\bibliography{acmart} 

\end{document}